\documentclass[graybox]{styles/svmult}

\usepackage{mathptmx}       
\usepackage{helvet}         
\usepackage{courier}        
\usepackage{type1cm}        
\usepackage{makeidx}         
\usepackage{graphicx} 
\usepackage{multicol}        
\usepackage[bottom]{footmisc}
\usepackage[numbers]{natbib}
\usepackage{subcaption}
\usepackage{algorithm}
\usepackage{amsfonts}
\usepackage{amsmath}
 \usepackage{threeparttable}

\newcommand{\Tab}[1]{Table~\ref{tab:#1}}

\usepackage{booktabs}
\usepackage{multirow}
\usepackage{hyperref}
\usepackage{diagbox}
\usepackage{xcolor}

\makeindex             

\begin{document}
\title*{Enhancing Personality Recognition in Dialogue by Data Augmentation and Heterogeneous Conversational Graph Networks}
\titlerunning{Enhancing Personality Recognition in Dialogue by Data Augmentation and HCGNN}
\author{
Yahui Fu, 
Haiyue Song, 
Tianyu Zhao,
Tatsuya Kawahara}
\institute{Yahui Fu, Tatsuya Kawahara \at Graduate School of Informatics, Kyoto University, Japan \\ \email{[fu][kawahara]@sap.ist.i.kyoto-u.ac.jp}
\and Haiyue Song\at Graduate School of Informatics, Kyoto University, Japan and NICT, Japan  \\  \email{song@nlp.ist.i.kyoto-u.ac.jp}
\and Tianyu Zhao\at rinna Co., Ltd., Japan \\ \email{tianyuz@rinna.co.jp}}
\maketitle

\renewcommand{\thefootnote}{\fnsymbol{footnote}}
\renewcommand{\thefootnote}{\arabic{footnote}}

\abstract{
Personality recognition is useful for enhancing robots' ability to tailor user-adaptive responses, thus fostering rich human-robot interactions. One of the challenges in this task is a limited number of speakers in existing dialogue corpora, which hampers the development of robust, speaker-independent personality recognition models.
Additionally, accurately modeling both the interdependencies among interlocutors and the intra-dependencies within the speaker in dialogues remains a significant issue.
To address the first challenge, we introduce personality trait interpolation for speaker data augmentation. For the second, we propose heterogeneous conversational graph networks to independently capture both contextual influences and inherent personality traits. Evaluations on the RealPersonaChat corpus demonstrate our method's significant improvements over existing baselines.\footnote{Our codes are available at \href{https://github.com/fuyahuii/Personality-Recognition-on-RealPersonaChat}{https://github.com/fuyahuii/Personality-Recognition-on-RealPersonaChat}.}
}

\begin{figure*}[t]
    \centering
    \includegraphics[width=0.8\linewidth]{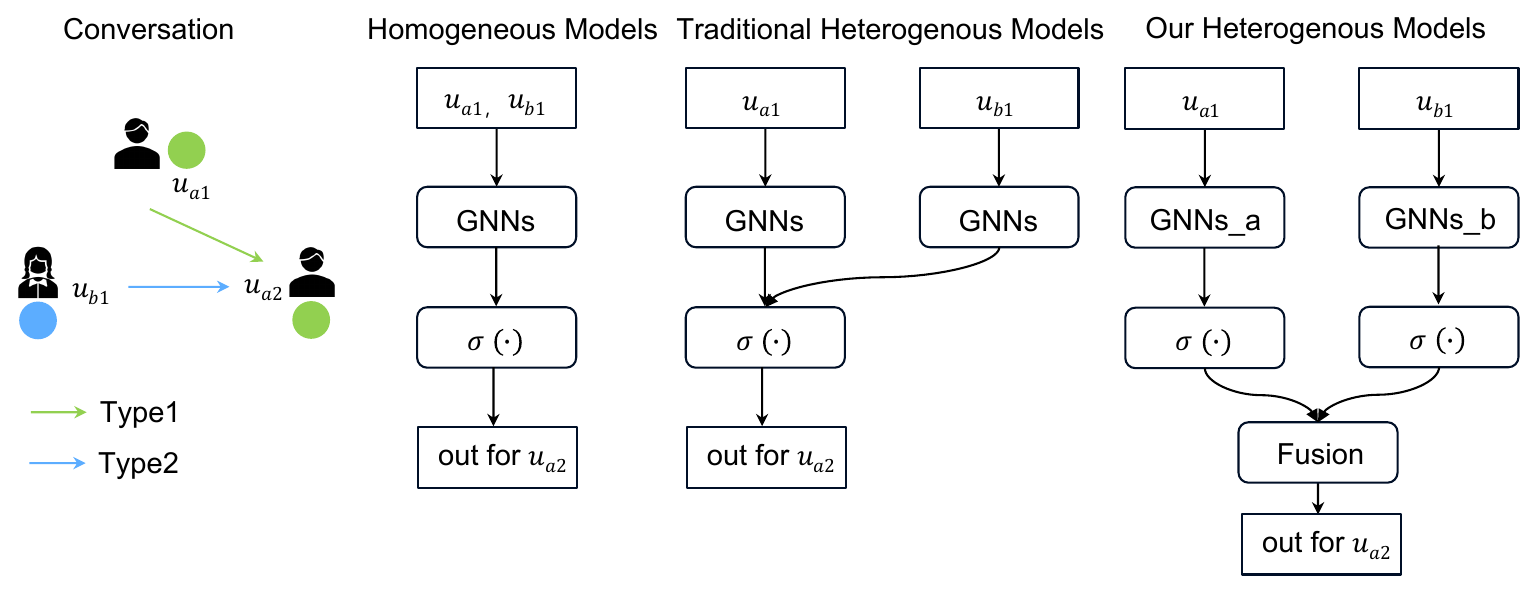}
    \caption{Homogeneous and different heterogeneous models. $u_{a1},u_{a2},u_{b1}$ represents alternant utterance of \textit{speaker} $a$ and $b$. $\sigma (\cdot)$ represents activation function.}
    \label{fig:my_label1}
\end{figure*}  

\section{Introduction}
\label{sec:1}
Personality recognition aims to identify an individual's characteristic patterns of feeling, thinking, and behaving, which make each different from one another~\cite{mairesse2007using}. Such capability is essential in the realm of human-robot interaction, where correctly detecting the user's personality can significantly enhance the robot's ability to tailor user-adaptive responses, thus fostering richer and more effective human-robot dialogues. Big-five traits \cite{mccrae1992introduction}, which encompass the dimensions of Openness, Neuroticism, Extraversion, Agreeableness, and Conscientiousness, and their respective counter-traits are commonly used in the community for personality assessment.
In contrast to prior approaches that infer personality traits from self-reported essays \cite{pennebaker1999linguistic,ramezani2022text, wen2023desprompt}, first impressions \cite{ponce2016chalearn,amin2023can}, or social media activities \cite{gjurkovic2020pandora}, this study focuses on the extraction of personality traits from dialogue \cite{rissola2019personality,jiang2020automatic,chen2022cped,mehl2006personality, Yamashita2023b}. 

However,  the lack of data is a major obstacle because annotating dialogue-level data with personality information is expensive and time-consuming. Each dialogue involves two participants and personality traits are obtained through psychology questionnaires. Thus, we investigate a data augmentation approach. While previous data augmentation studies focus on generating sentence-level data \textit{invariants}~\cite{dhole2021nl,wei-zou-2019-eda,gao-etal-2020-paraphrase,sennrich-etal-2016-improving} without corresponding labels, in this study, we generate both the synthetic dialogue data and corresponding synthetic personality traits through the proposed data interpolation method, which fuses two existing data points controlled by a \textit{continuous ratio variable}.

Additionally, accurately modeling both the inter-dependencies between context and interlocutors, as well as the intra-dependencies within speakers in dialogues, remains a significant challenge. Previous homogeneous models, such as the graph attention network \cite{velivckovic2017graph} \cite{brody2021attentive},
did not consider the variations in link types. Heterogeneous models like relational graph convolution networks (RGCN) employ distinct relation types to model various dependencies. Yet, they utilize shared coefficients across all relation types, which may fail to capture the unique attributes of each relation type, as shown in Fig~\ref{fig:my_label1}. To address this issue, we propose a method to independently model heterogeneous conversational interactions, capturing both contextual influences and inherent personality traits. 
Our main contributions are as follows:
\begin{itemize}
    \item{We propose a data augmentation method for personality recognition by interpolation from any two existing data points.}
    \item{We propose a heterogeneous conversational graph neural network (HC-GNN) to independently model both the interdependencies among interlocutors, as well as the intra-dependencies within the speaker in dialogues.}
    \item{Experimental results using the RealPersonaChat corpus demonstrate that increasing speaker diversity significantly improves personality recognition in both monologue and dialogue settings. The proposed HC-GNN method outperforms baseline models, showcasing its effectiveness.}
\end{itemize}


%



\section{Related Work}
\label{sec:2}
\subsection{Personality Recognition in Dialogue}
Mehl et al. \cite{mehl2006personality} pioneered the automatic personality assessment of all Big Five personality traits using various psycholinguistic attributes. To this end, they analyzed a collection of daily-life conversations by 96 participants over 2 days. 
However, it only contains the subjects' conversation; we also want to analyze how interlocutors impact the subject's personality expression in the dialogue.
Jiang et al. \cite{jiang2020automatic} and Chen et al. \cite{chen2022cped} collected conversation corpus with rich data based on the ``Friends" and 40 Chinese television series, respectively. 
Nonetheless, the Big Five personality labels were assigned by external observers rather than derived from self-assessments by the speakers. 
Most recently, Yamashita et al. \cite{Yamashita2023b} presented the RealPersonaChat (RPC) corpus by documenting the authentic personality traits of the participants and allowing them to freely engage in dialogues. This corpus aligns closely with our research objectives, as it provides a foundation for evaluating the personality traits of subjects who may engage in chit-chatting dialogue with a conversational agent. However, this corpus has a relatively limited number of speakers (233). This sparsity poses a challenge in effectively detecting the personality traits of unseen speakers. We propose a data augmentation method to enrich the speaker diversity.

\subsection{Data Augmentation} 
Data augmentation (DA) tries to fill the gap between the data distribution of the training set and the real data with no annotation cost.
Previous DA studies focus on generating data \textit{invariants}. In the computer vision field, simple geometric transformations like cropping, rotation, and noise injection can be easily applied to continuous image data~\cite{shorten2019survey}. Due to the discrete nature of language data, previous DA studies in NLP usually involve discrete noises including \textit{1)} character-level modification like changing character case~\cite{dhole2021nl}, \textit{2)} subword-level regularization such as BPE-dropout~\cite{provilkov-etal-2020-bpe}, \textit{3)} word-level replacement, insertion, or deletion~\cite{wei-zou-2019-eda}, and \textit{4)} sentence-level modification such as paraphrasing~\cite{wang2019task,gao-etal-2020-paraphrase} and back-translation~\cite{sennrich-etal-2016-improving,xie-etal-2018-noising,graca-etal-2019-generalizing,mojapelo2023data}.
Our method differs from them in two aspects. First, we generate data \textit{variants} including both dialogue data and corresponding personality labels.
Second, we generate synthetic personality traits following continuous distribution from existing discrete trait data by introducing a random fusion variable.
This bridges the gap between the discrete distribution in the corpus and the continuous distribution in reality.
A similar work is the example extrapolation method~\cite{wei-2021-good} which generates augmented embeddings of the target domain by leveraging the similarity of embedding spaces from another assisting domain. Different from it, our interpolation method requires only one dataset.

\subsection{Graph Neural Networks}
\label{subsec:2} 
Graph Neural Networks (GNNs) and their variants have been widely applied in dialogue-related tasks, like conversational emotion recognition \cite{fu2021consk,fu2022context}, and dialog act classification \cite{fu2023hag}. This is primarily due to the adjacency matrix in GNNs effectively simulating interactions within conversations.
Yang et al. \cite{yang2023orders} proposed a dynamic deep graph convolutional network for personality detection on social media posts. 
Existing methodologies, whether employing static or dynamic approaches to construct interactions within graphs, mainly focus on homogeneous or heterogeneous conversation modeling. Nevertheless, various types of nodes and links have different traits and their features may fall in different spaces. For instance, as illustrated in Fig~\ref{fig:my_label1}, traditional heterogeneous models like RGCN \cite{schlichtkrull2018modeling} utilize shared coefficients across all relation types, potentially failing to capture the unique attributes of each relation type.
To solve this, this paper proposes a modification to the existing heterogeneous model framework. We introduce separate GNNs to distinctly capture the diverse relation types, thereby respecting the unique properties of each node and link type.

\section{Preliminary Dataset Analysis}
\begin{table}[htbp]
\caption{Pearson correlation between pairs of Big-Five personality traits in the dataset. N, E, O, A, and C represent Neuroticism, Extraversion, Openness, Agreeableness, and Conscientiousness, respectively.}
\label{tab:my-table1}
\centering
\small
\begin{tabular}{@{}lcccccccccc@{}}
\toprule
Big-Five Pair & (N, E) & (N, O) & (N, A) & (N, C) & (E, O) & (E, A) & (E, C) & (O, A) & (O, C) & (A, C) \\ \midrule
Pearson Correlation & -0.49  & -0.23  & -0.27  & -0.15  & 0.47   & 0.46   & 0.26   & 0.40   & 0.15   & 0.36   \\ \bottomrule
&&&&&&\multicolumn{5}{r}{(for all pairs $p<.05$)}\\
\end{tabular}
\end{table}

We conduct experiments using the RealPersonaChat corpus \cite{Yamashita2023b}, comprising 14,000 Japanese dialogues and a total of 421,203 utterances. In this corpus, 233 participants (141 females, 90 males) completed a questionnaire regarding their Big Five personality score (in a range from 1-7) and then engaged in unstructured conversation. We normalize the score to 0-1. We analyze the correlations between pairs of personality traits in the dataset, as shown in Table~\ref{tab:my-table1}. The $p$-value of the Pearson correlation for each pair is less than 0.05, indicating statistically significant relationships between each pair of Big-Five personality traits.
    
\section{Proposed Method}

\subsection{Data Interpolation}
\label{method_data_interpolation}
This section describes math notations, how to fuse two existing data points to generate synthetic dialogue and Big-five traits, and variants of the proposed method.


\noindent \textbf{Notations.} Each dialogue $D$ contains utterances $u_{a}$ from the \textit{speaker a} or $u_{b}$ from the \textit{speaker b} in alternant turns, that is $D=\{u_{a1},u_{b1},u_{a2},u_{b2},...,u_{an},u_{bn}\}$, where $n$ is the number of turns. Each dialogue is accompanied by a label $\textbf{y}$ that is a vector containing the Big-five personality traits of the target speaker. 
We aim to generate synthetic dialogue $D_{syn}$ and its label $\textbf{y}_{syn}$ from two existing dialogues $D_1$ and $D_2$ and their labels $\textbf{y}_1$ and $\textbf{y}_2$ 
.\footnote{Since we focus on the personality of the initiating \textit{speaker a} in the experiments, $\textbf{y}_1$ or $\textbf{y}_2$ refers to the personality of \textit{speaker a}.}

\noindent \textbf{Dialogue Interpolation.} 
First, we randomly select two dialogues $D_1$ and $D_2$ in the training set. 
Second, we split each dialogue into chunks ($c$) each containing $t$ turns, which is a hyper-parameter controlling the context length we desire (we set $t$=3). 
This results in $D_1=\{c_1, c_2, ..., c_{l}\}$ and $D_2=\{c^{\prime}_1, c^{\prime}_2, ..., c^{\prime}_{l}\}$, where $l=\frac{n}{t}$ is the number of chunk one dialogue contains. 
Finally, we combine chunks from $D_1$ and $D_2$ to generate $D_{syn}$ using a fusion ratio $\beta$ that is a random variable independently sampled from $\text{Uniform}(0, 1)$ for each synthetic data point. $D_{syn}$ can be represented as:
\begin{equation}
    \begin{aligned}
        D_{\text{syn}} &= \beta D_1 \oplus (1-\beta)D_2, \text{resulting in} \\
        D_{\text{syn}} &= \{c^{\text{syn}}_i \mid 1 \leq i \leq l\}, \text{where} \\
        c^{\text{syn}}_i &= 
        \begin{cases} 
            c_i & \text{with probability } \beta, \\
            c^{\prime}_i & \text{with probability } 1-\beta.
        \end{cases}
    \end{aligned}
\end{equation}
Specially, when generating synthetic monologue data, we split each monologue $D$ into utterances instead of chunks.

\noindent \textbf{Label Interpolation.} Because each label is a vector of real numbers represented as $\textbf{y} \in \mathbb{R}^5$, we can simply obtain the synthetic label through:
\begin{equation}
    \begin{aligned}
     \textbf{y}_{syn}=\beta \textbf{y}_1+(1-\beta)\textbf{y}_2.
    \end{aligned}
\end{equation}

\noindent \textbf{Method Variants.} 
There are three types of variants of the proposed method (former setting used).
First, we can sample $\beta \sim \text{Uniform}(0, 1)$ or fixing $\beta$ to $0.5$. The former setting produces a richer variety of data. 
Second, we can either select two dialogues possibly from different speakers, where $\textbf{y}_1$ and $\textbf{y}_2$ are independently and identically distributed ($\textbf{y}_1 \overset{\text{iid}}{\sim} \textbf{y}_2$), or select two dialogues from the same speaker $\textbf{y}_1=\textbf{y}_2$ which results in $\textbf{y}_1=\textbf{y}_2=\textbf{y}_{syn}$. The former setting can produce new speakers with synthetic personality traits.
Third, the synthetic dialogue can be truncated to length  $l^\prime \sim \text{Uniform}(t_{min}, |D_{syn}|)$ (we set $t_{min}=2$).
This enables personality recognition in early turns which is preferred in real application.

\subsection{Heterogeneous Conversational Graph Neural Network}

\begin{figure*}[t]
    \centering
    \includegraphics[width=\linewidth]{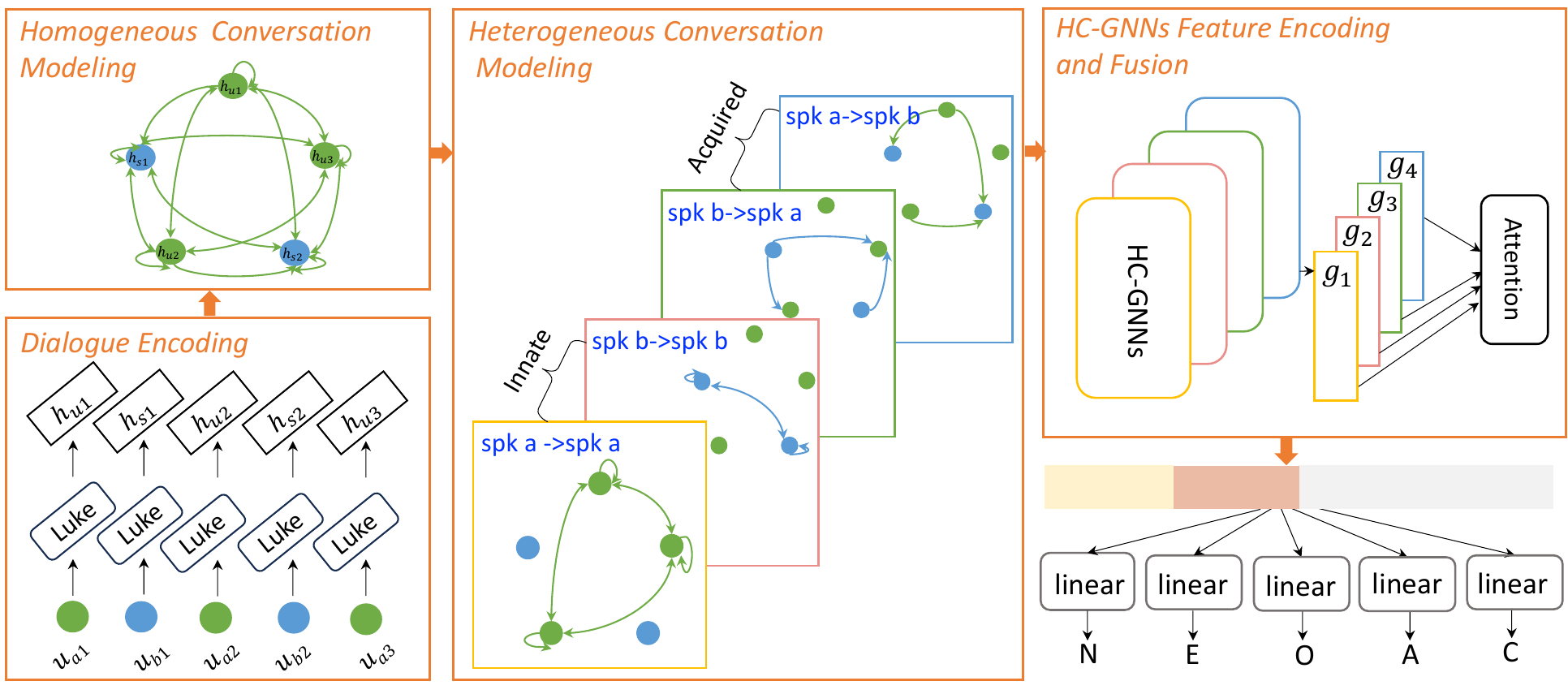}
    \caption{\textbf{Heterogeneous conversational graph neural network (HC-GNN)}, which captures the interdependencies among interlocutors (acquired) and the intra-dependencies within \textit{speaker a} or \textit{b} (innate).}
    \label{fig:my_label2}
\end{figure*}

We map each utterance in the dialogue into embeddings, as presented in Section~\ref{DialogueEncoding}. Subsequently, we describe heterogeneous conversation modeling in Section~\ref{HeterogeneousConversationModeling}, followed by an explanation of heterogeneous conversational graph neural network feature encoding and fusion in Section~\ref{HCGCNNEncoding}.

\subsubsection{Dialogue Encoding}
\label{DialogueEncoding}

For each dialogue $D=\{u_{a1},u_{b1},u_{a2},u_{b2},u_{a3}...\}$, we employ a BERT-like model Japanese Language Understanding with
Knowledge-based Embeddings (LUKE) \cite{yamada2020luke} \footnote{\href{https://huggingface.co/studio-ousia/luke-japanese-base}{https://huggingface.co/studio-ousia/luke-japanese-base}} to encode each utterance in the dialogue:
\begin{equation}
\textbf{h}_{u_{i}}=\texttt{LUKE} (u_{i}) \in \mathbb{R}^{1 \times d}
\end{equation}
where $h_{u_{i}}$ denotes the final hidden state of the ``[CLS]" token to represent the meaning of the whole utterance, and $d$ is the dimension of the output.

\subsubsection{Heterogeneous Conversation Modeling}
\label{HeterogeneousConversationModeling}
To explicitly model the interaction between speakers, we independently model the intra-dependency (innate personality) and inter-dependency (acquired personality which is influenced by the interlocutor), as shown in Fig.~\ref{fig:my_label2}.
We introduce the following notation: we denote directed and labeled multi-graphs as $\mathcal{G}_{r} = (\mathcal{V}_{r}, \mathcal{E}_{r}, r)$ with nodes $\vartheta_{r, i} \in \mathcal{V}_{r}$ and labeled edges (relations) $(\vartheta_{r, i},r, \vartheta_{r,j}) \in \mathcal{E}_{r}$, where 
$r \in \mathcal{R}$ represents one of the conversation relation types \{\textit{$spk$ $a$ $\rightarrow$ $spk$ $a$}, \textit{$spk$ $b$ $\rightarrow$ $spk$ $b$}, \textit{$spk$ $b$ $\rightarrow$ $spk$ $a$}, \textit{$spk$ $a$ $\rightarrow$ $spk$ $b$}\}
.

\subsubsection{HC-GNNs Feature Encoding and Fusion}
\label{HCGCNNEncoding}
For each relation type in each graph $\mathcal{G}$, we then encode the features of node $i$ with dynamic attention and graph attention networks (GATv2) \cite{brody2021attentive} to aggregate the interactions between each group of speakers (self and interlocutor):

\begin{equation}
  \begin{aligned}
    &\textbf{h}_{i}^{n(l)}=\sigma (\sum_{k\in K }\sum_{j\in N_{i}^{r}}\frac{a_{i,j}^{(k)}}{N_{i}^{r}}\textbf{W}_{r}^{(l)}\textbf{h}_{u_{j}}+a_{i,i}^{(k)}\textbf{W}_{0}^{(l)}\textbf{h}_{u_{i}})  \\
  \end{aligned}  
\end{equation}
where $N_{i}^{r}$ denotes the neighboring indices of node $i$ under relation $r\in\mathcal{R}$.
This notation encompasses both forward and backward relation directions.
$K$ represents the number of attention head. $\textbf{W}_{r}^{(l)}$, $\textbf{W}_{0}^{(l)}$ are the learnable weight metrics, $l$ is the number of layer of HC-GNN, and $\sigma(.)$ is an ReLU activation function. 
The attention scores are normalized across all neighbors $j \in N_{i}^{r}$ using softmax, and the attention function is defined as:
\begin{equation}
  \begin{aligned}
 &a_{i,j}=softmax_{j} (e(\textbf{h}_{u_{i}},\textbf{h}_{u_{j}}))=\frac{exp(e(\textbf{h}_{u_{i}},\textbf{h}_{u_{j}}))}{\sum_{j^{'}\in N_{i}^{r}}exp(e(\textbf{h}_{u_{i}},\textbf{h}_{u_{j'}}))}\\
    &e(\textbf{h}_{u_{i}},\textbf{h}_{u_{j}})=\mathbf a^{\top }LeakyReLU(\textbf{W} \cdot  [\textbf{h}_{u_{i}}||\textbf{h}_{u_{j}}])
  \end{aligned}  
\end{equation}
where $\mathbf a\in {\mathbb{R}^{d}}$ and $\textbf{W}\in {\mathbb{R}^{d^{'}\times d}}$ are learned, $||$ denotes vector concatenation. We use graph neural network (GCN) \cite{velivckovic2017graph} to capture the deeper interaction representations:
\begin{equation}
    \begin{aligned}
       &\textbf{h}_{i}^{n(l+1)}=\sigma (\sum_{j\in N_{i}^{r}}\textbf{W}^{(l+1)}\textbf{h}_{j}^{n(l)}+\textbf{W}_{0}^{(l+1)}\textbf{h}_{i}^{n(l)})\\
    \end{aligned}
\end{equation}
where $\textbf{W}^{(l+1)}$,$\textbf{W}_{0}^{(l+1)}$ are learnable metrics. Given the latent representation $\textbf{g}_{r}$ \cite{velivckovic2017graph} of each graph which corresponds to a distinct relation, we then use the self-attention mechanism to fuse the graph outputs of innate and acquired relations:
\begin{equation}
\begin{aligned}
&\textbf{z}=\textbf{g}_{0}||\textbf{g}_{1}||...||\textbf{g}_{r}\\
    &\textbf{z}^{'}=Attn (\textbf{z,z,z})
\end{aligned}
\end{equation}

\subsection{Personality Recognition with Multi-task Learning}
\label{PersonalityRecognition}
Since each pair of Big-Five is statistically significantly related as shown in Table~\ref{tab:my-table1}, we recognize Openness, Neuroticism, Extraversion, Agreeableness, and Conscientiousness in the manner of multi-task learning using five linear layers. The $i_{th}$ layer can be represented as:
\begin{equation}
    \mathcal{P}_{i}=\sigma(\textbf{W}_{p}^{i}\textbf{z}^{'}+b_{p}^{i})
\end{equation}
where $\sigma$ denotes the activation function ReLU, $\textbf{W}_{p}^{i}$ and $b_{p}^{i}$ are the learnable weight matrix and bias. 
We treat them as regression tasks and use the mean absolute error (MAE) as the loss function for model optimization. The loss item of each data sample $j$ is denoted as $l(\mathcal{\textbf{P}}^{j}, \textbf{y}^{j})$ that is calculated by averaging the loss of five tasks, where $\mathcal{\textbf{P}}^{j}$ is the vector of $5$ predictions and $\textbf{y}^{j}$ is the vector of $5$ ground truth personality traits.
The loss item of one batch containing $N$ data samples is denoted as $L$. They are calculated as follows:
\begin{equation}
\begin{aligned}
    &l(\mathcal{\textbf{P}}^{j}, \textbf{y}^{j}) = \frac{1}{5} \sum_{i=1}^{5} \left| \mathcal{P}^{j}_{i} - y^{j}_i \right| \\
    & L=\frac{1}{N}\sum_{j=1}^{N} l(\mathcal{\textbf{P}}^{j}, \textbf{y}^{j}).
\end{aligned}
\end{equation}

\section{Experimental Settings}
\subsection{Dataset Partitioning}
We conduct two experimental settings, one based on \textbf{monologues} and the other on \textbf{dialogues}. The monologue setting focuses on a speaker's own utterances, while the dialogue setting integrates utterances from both the speaker and the interlocutor for personality recognition. In the monologue experiments, we implemented strict speaker splitting to ensure no overlap among speakers across the training, validation, and test sets. This approach meant the model was evaluated on unseen speakers. 

In the dialogue experiments, ensuring non-overlap of both speakers in all datasets proved challenging. Therefore, we ensured that only the initiating speaker was non-repeating across datasets, and the model was tasked with predicting only the initiating speaker's personality. Therefore, we only use relation \textit{spk a$\rightarrow$spk a} and \textit{spk b$\rightarrow$spk a} in the HC-GNN model.
We created training, validation, and test sets by randomly dividing the speakers in an 8:1:1 ratio for $100$ times and selecting the split that most closely matched 8:1:1 distribution for monologues and dialogues. We then fixed the split across all experiments. 


%
%


\subsection{Models}

All the models are based on a BERT-like model which converts utterances into embeddings. In all experiments, we used pooled output from LUKE base model~\cite{yamada2020luke} because it showed the best performance in terms of average balanced accuracy among LUKE-base/large, RoBERTa-base/large, xlm-roberta base/large, mdeberta-v3-base models.
We experimented with different models after the base model:

\noindent\textbf{MLP}: Five linear module joints with regression heads are used to predict each personality. Each regression head contains $2$ linear layers where the first layer maps embedding from LUKE to embedding with a size $16$ and the second layer maps embedding with a size $16$ to $1$. 

\noindent\textbf{Homogeneous Methods}:

\noindent \textbf{GCNs} \cite{kipf2016semi}: represents one of the most prevalent methods for handling graph-structured data, particularly in node classification and link prediction tasks.
    
\noindent \textbf{GATv2} \cite{brody2021attentive}:
introduces a significant enhancement by transitioning from a static to a dynamic attention mechanism. 

\noindent In our experiments, we utilize either two GraphConv layers or a single GATv2Conv layer to process the union of relation types within the set $\mathcal{R}$. 

\noindent\textbf{Heterogenous Methods}:

\noindent \textbf{RGCN} \cite{schlichtkrull2018modeling}: relational graph convolutional network, which is developed to handle multi-relational data with a heterogeneous architecture, as shown in Fig.~\ref{fig:my_label1}. Specifically, we use one RGCNConv layer connected by one GraphConv layer to handle the union of relation types in $\mathcal{R}$, setting the number of relations to 2.

\noindent \textbf{HCGNN}: our proposed method, the heterogeneous conversation graph neural network, independently models the interdependencies and intra-dependencies within speakers. Specifically, a GATv2Conv layer, interconnected with two GraphConv layers, is employed to extract distinct relational information corresponding to each relation type in the set $\mathcal{R}$.
\noindent GraphConv, GATv2Conv and RGCNConv layers are imported from torch\_geometric.nn \footnote{
\href{https://pytorch-geometric.readthedocs.io/en/latest/modules/nn.html}{https://pytorch-geometric.readthedocs.io/en/latest/modules/nn.html}}.
The number of the attention head in GATv2Conv is set to 2.

\subsection{Training}

We used the Adam optimizer with \(\beta_1 = 0.9, \beta_2 = 0.999\) and a learning rate of \(1 \times 10^{-5}\). We used a linear scheduler with warmup step $= 150$. We used Mean Absolute Error (MAE) criterion because it outperformed the Mean Squared Error (MSE) greatly in terms of balanced accuracy.  We set the batch size to $128$ for linear models and $32$ for graph neural network models, which reach the memory limitation of eight 32G GPUs.
We calculated the loss on the validation set after each epoch and applied early stopping when no improvement was observed for $3$ epochs. 
We conducted single-round experiments for each configuration. We employed four distinct metrics to assess the model's performance across five different personality traits, which will offer insight into the model’s overall capabilities.

\subsection{Evaluation Metrics}
We report binary classification accuracy and balanced accuracy together with Pearson correlation and Spearman correlation for regression tasks.
The threshold for the binary classification task of each personality trait is set to the median score in the training set. Here are the details of each metric:

\noindent\textbf{Accuracy}: a metric that summarizes the performance of a classification task, which is the number of correctly predicted data points out of all the data points.

\noindent\textbf{Balanced Accuracy}: arithmetic mean of sensitivity and specificity to deal with imbalanced data.

\noindent\textbf{Pearson Correlation}: a correlation coefficient that measures the linear correlation between the predicted personality values and the ground truth.

\noindent\textbf{Spearman Correlation}: a nonparametric measure of rank correlation (statistical dependence between the rankings of two variables).

\section{Results and Analysis}
\subsection{Monologue}
\label{sec:monologue_res}

\begin{table*}[thb]
\caption{Accuracy results in monologue setting with original data and augmented dataset. 
The best result in each column is in \textbf{bold}.}
\label{tab:mono_acc}
\centering
\small
\begin{tabular}{@{}lcccccc|cccccc@{}}
\toprule
\multirow{2}{*}{Data}& \multicolumn{6}{c|}{Accuracy} & \multicolumn{6}{c}{Balanced Accuracy}\\
\cmidrule(l){2-13} 
& N     & E     & O     & A     & C     & Avg. & N     & E     & O     & A     & C     & Avg. \\ \cmidrule(l){2-13} 
Original&66.2&52.6&57.6&52.7&57.9&57.4&59.7&54.3&58.8&50.3&55.0&55.6\\
+10k&74.3&55.5&56.6&49.9&51.2&57.5&60.4&52.5&55.4&47.8&52.8&53.8\\
+20k&\textbf{74.5}&54.3&55.5&\textbf{59.0}&60.1&60.7&\textbf{65.9}&53.8&56.0&58.5&48.3&56.5\\
+50k&60.5&53.0&60.9&58.7&59.8&58.6&60.2&55.0&61.2&\textbf{62.0}&52.4&58.2\\
+500k&62.7&\textbf{58.2}&\textbf{62.0}&57.9&\textbf{65.4}&\textbf{61.2}&64.6&\textbf{56.0}&\textbf{61.3}&60.3&\textbf{59.7}&\textbf{60.4}\\\bottomrule
\end{tabular}
\end{table*}

\begin{table}[thb]
\caption{Correlation results in monologue setting with various data sizes.}
\centering
\label{tab:mono_correlation}
\small
\begin{tabular}{@{}lrrrrrrrrrr@{}}
\toprule
\multirow{2}{*}{Data} & \multicolumn{5}{c|}{Pearson Correlation} & \multicolumn{5}{c}{Spearman Correlation}\\
\cmidrule(l){2-11} 
& \multicolumn{1}{c}{N} & \multicolumn{1}{c}{E} & \multicolumn{1}{c}{O} & \multicolumn{1}{c}{A} & \multicolumn{1}{c|}{C}     & \multicolumn{1}{c}{N} & \multicolumn{1}{c}{E} & \multicolumn{1}{c}{O} & \multicolumn{1}{c}{A} & \multicolumn{1}{c}{C} \\ 
\cmidrule(l){2-11} 
Original                         & \textit{.279} & -.040 & \textit{.473} & \textit{.105} & \multicolumn{1}{r|}{\textit{.166}}  & \textit{.292} & -.014 & \textit{.282} & \textit{.104} & \textit{.200} \\
+10k                             & \textit{.389} & -.092 & \textit{.510} & \textit{.100} & \multicolumn{1}{r|}{\textbf{\textit{.200}}} & \textit{.283} & -.064 & \textit{.277} & \textit{.091} & \textit{.180} \\
+20k                             & \textbf{\textit{.492}} & .054  & \textbf{\textit{.510}} & \textit{.268} & \multicolumn{1}{r|}{\textit{.113}}  & \textbf{\textit{.495}} & .063  & \textit{.276} & \textit{.261} & \textit{.126} \\
+50k                             & \textit{.224} & .067  & \textit{.486} & \textbf{\textit{.387}} & \multicolumn{1}{r|}{\textit{.099}}  & \textit{.264} & .085  & \textit{.378} & \textbf{\textit{.375}} & \textit{.119} \\
+500k                            & \textit{.267} & .025  & \textit{.459} & \textit{.232} & \multicolumn{1}{r|}{\textit{.164}}  & \textit{.333} & .045  & \textbf{\textit{.388}} & \textit{.230} & \textbf{\textit{.203}} \\ \bottomrule
&&&&&&\multicolumn{5}{r}{(\textit{Italic} means $p<.05$)}
\end{tabular}
\end{table}

\noindent \textbf{Main Results.} 
\Tab{mono_acc} presents a comparative analysis of the accuracy and balanced accuracy of the data augmentation method at various data sizes. 
With the addition of augmented data, both accuracy and balanced accuracy show significant improvements, increasing from 57.4\% to 61.2\% and from 55.6\% to 60.4\%, respectively.
    \Tab{mono_correlation} shows the Pearson and Spearman correlation results. We observe that data augmentation generally improves correlation in most traits, and the impact of augmentation is more pronounced in N and A than in others. We fail to predict personality trait E in any setting. We believe this may be due to our dataset being based on first-meeting spontaneous situations, where people tend not to exhibit extrovert traits explicitly.
We also find that the optimal results for different personality traits, in terms of accuracy, balanced accuracy, Pearson correlation, and Spearman correlation, do not consistently align with the same data augmentation size. We attribute this variability primarily to the inherent randomness associated with data synthesis.

\noindent \textbf{Results of Data Augmentation Variants.}

\noindent \textit{1. Fusing Ratio.} We compare results using $\beta \sim \text{Uniform}(0, 1)$ and fixed $\beta=0.5$. With 500k additional data, using random $\beta$ achieved $61.2\%$ averaged accuracy and $60.4\%$ averaged balanced accuracy whereas using fixed $\beta$ showed $59.3\%$ accuracy and $58.4\%$ balanced accuracy. This is because using random $\beta$ results in more various data.
Since the original label score distribution is discrete (there are only a limited number of score values), using $\beta$ following a continuous distribution yields a higher variety of labels and corresponding textual data compared to using fixed $\beta$.

\begin{figure*}[thb]
    \centering
    \begin{subfigure}{\textwidth}
        \centering
        \includegraphics[width=\textwidth]{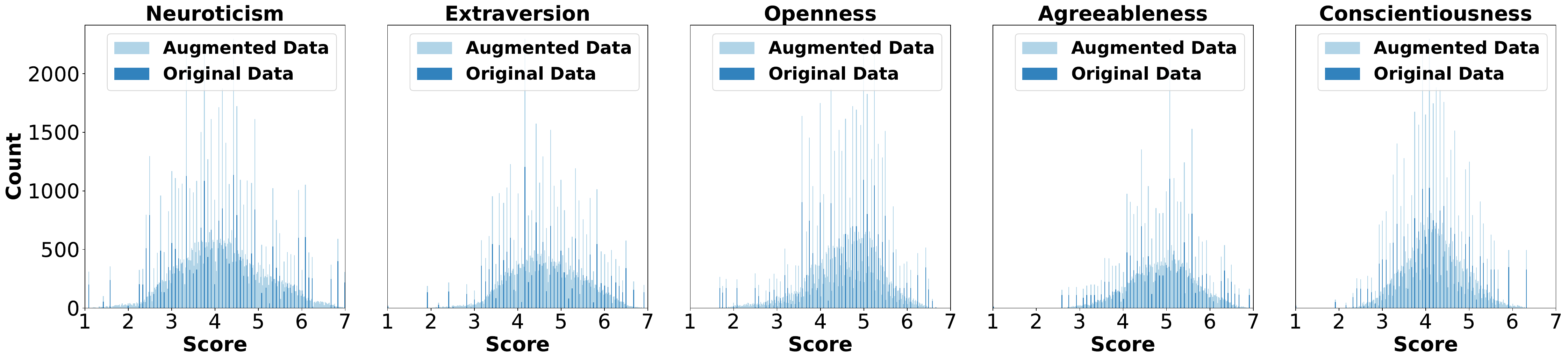}
        \caption{Generate synthetic data by combining two dialogues from two different speakers.}
        \label{fig:interpolation_res_a}
    \end{subfigure}
    \begin{subfigure}{\textwidth}
        \centering
        \includegraphics[width=\textwidth]{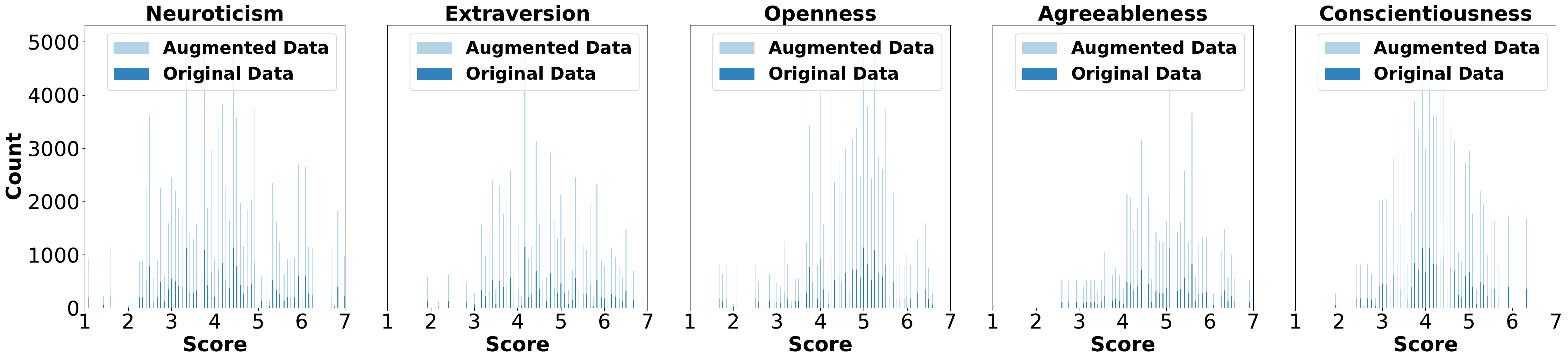}
        \caption{Generate synthetic data by combining two dialogues from the same speaker.}
        \label{fig:interpolation_res_b}
    \end{subfigure}
    \caption{Data distribution of augmented data and original data.}
    \label{fig:interpolation_res}
\end{figure*}

\noindent \textit{2. Speaker Choice.} We compare generating synthetic dialogue from the same speaker or two different speakers. We observe using dialogues from different speakers not only enables continuous data distribution as shown in Fig.~\ref{fig:interpolation_res} but also showed much higher averaged accuracy ($61.2\%$ vs $57.3\%$) and balanced accuracy ($60.4\%$ vs $58.1\%$), which demonstrates that speaker variety is more crucial than the number of conversations for personality recognition within this dataset.

\noindent \textit{3. Various Context Lengths.} Real-time personality recognition in dialogue is essential for human-robot interaction. To enable this, we propose to use various context lengths of dialogues for training.
\Tab{various_length_res} presents comparative results on a range of test context turns when training with various context lengths and full context lengths, using an augmented data set of 500k. When testing the first $2$ turns using various context lengths during training results in $58.2\%$ averaged balanced accuracy, which is comparable to the result of $60.4\%$ using the full contexts during inference. On the other hand, we got only $55.5\%$ when training with full contexts (approximately $15$ turns).
In addition to context turns being 2 during inference, similar trends can also be observed in other numbers.

\vspace{-0.7cm}
\begin{table}[thb]
\caption{Comparative results of accuracy and balanced accuracy for a range of test context turns: training with various context turns ($\geq2$) vs. full context turns, using an augmented data set of 500k.}
\centering
\label{tab:various_length_res}
\small
\begin{tabular}{lcc|cc}
\toprule
\multirow{2}{*}{\diagbox{test}{train}} & \multicolumn{2}{c|}{Accuracy} & \multicolumn{2}{c}{Balanced Accuracy} \\
& various context & full context & various context & full context \\
\midrule
2&59.7&57.8&58.2&55.5\\
3&59.6&57.6&57.8&55.8\\
4&59.9&58.1&58.4&56.6\\
5&60.3&58.0&58.6&56.3\\
10&60.7&57.6&59.6&56.6\\
full&61.2&57.1&60.4&57.0\\
\bottomrule
\end{tabular}
\end{table}

\begin{table}[thb]
\caption{Accuracy results in the comparisons among monologue and dialogue.}
\label{tab:accu_mono_dialog}
\centering
\scalebox{0.93}{
\begin{tabular}{@{}llcccccc|cccccc@{}}
\toprule
\multicolumn{2}{c}{\multirow{2}{*}{\;\;\;\,Model}}      & \multicolumn{6}{c|}{Accuracy}      & \multicolumn{6}{c}{Balanced Accuracy}               \\ \cmidrule(l){3-14} 
\multicolumn{2}{l}{}           & N     & E     & O     & A     & C     & Avg. & N     & E     & O     & A     & C     & Avg. \\ \midrule
Monologue&MLP&66.2&52.6&57.6&52.7&57.9&57.4&59.7&54.3&\textbf{58.8}&50.3&\textbf{55.0}&55.6\\\midrule
\multirow{6}{*}{Dialogue} & MLP&66.8&53.2&\textbf{57.9}&49.0&58.4&57.1&59.5&\textbf{56.9}&57.9&49.8&47.5&54.3 \\
                          &GCN~\cite{kipf2016semi}&\textbf{69.5}&40.3&52.9&33.3&\textbf{72.1}&53.6&50.0&50.4&53.7&34.4&50.1&47.7\\
 &GAT~ \cite{brody2021attentive}&66.4&54.5&52.7&59.9&54.6&57.6&60.4&51.3&50.1&52.2&52.1&53.2\\ &RGCN~\cite{schlichtkrull2018modeling}&68.5&\textbf{57.2}&55.4&41.9&56.7&55.9&\textbf{63.0}&54.5&55.9&47.2&51.9&54.5\\
 \cmidrule(l){2-14} 
 &HC-GNN (ours) &69.0&53.5&55.6&\textbf{65.9}&52.2&\textbf{59.2}&60.9&54.3&52.6&\textbf{59.6}&54.6&\textbf{56.4}\\ \bottomrule
\end{tabular}}
\end{table}

\begin{table}[htb]
\caption{Correlation results in the comparisons among monologue and dialogue.}
\centering
\label{tab:corre_mono_dialog}
\begin{tabular}{@{}llccccc|ccccc@{}}
\toprule
\multicolumn{2}{c}{\multirow{2}{*}{\;\;\;\,Model}}      & \multicolumn{5}{c|}{Pearson Correlation}      & \multicolumn{5}{c}{Spearman Correlation}               \\ \cmidrule(l){3-12} 
\multicolumn{2}{l}{}           & N     & E     & O     & A     & C     & N     & E     & O     & A     & C     \\ \midrule
Monologue & MLP & \textit{.279} & -.040 & \textit{.473} & \textit{.105} & \textit{.166} & \textit{.292} & -.014 & \textit{.282} & \textit{.104} & \textbf{\textit{.200}} \\ 
\midrule
\multirow{5}{*}{Dialogue}
& MLP & \textit{.288} & .048 & \textbf{\textit{.492}} & \textit{.174} & \textit{.077} & \textit{.214} & .088 & \textit{.262} & \textit{.123} & \textit{.072} \\
& GCN \cite{kipf2016semi} & \textit{.170} & -.015 & \textit{.298} & \textit{-.203} & \textit{.079} & \textit{.173} & .017 & \textit{.164} & \textit{-.227} & \textit{.105} \\
& GAT \cite{brody2021attentive} & \textit{.307} & -.067 & \textit{.420} & \textit{.079} & \textit{.134} & \textit{.284} & -.030 & \textit{.226} & \textit{.082} & \textit{.135} \\
& RGCN \cite{schlichtkrull2018modeling} & \textbf{\textit{.377}} & .048 & \textit{.490} & \textit{.152} & \textit{.148} & \textbf{\textit{.375}} & .078 & \textbf{\textit{.311}} & \textit{.146} & \textit{.160} \\
\cmidrule(l){2-12} 
& HC-GNN (ours) & \textit{.285} & .040 & \textit{.347} & \textbf{\textit{.216}} & \textbf{\textit{.169}} & \textit{.304} & .066 & \textit{.243} & \textbf{\textit{.191}} & \textit{.193} \\
\bottomrule
&&&&&&\multicolumn{6}{r}{(\textit{Italic} means $p<.05$)}
\end{tabular}
\end{table}

\vspace{-0.8cm}
\subsection{Dialogue}
\label{sec:dialogue_res}
\noindent\textbf{Comparisons between Monologue and Dialogue.} To explore the impact of context on personality recognition, we first prepend the [SPK1] or [SPK2] token to the respective utterances and then concatenate all utterances using the [SEP] token. We employ the same model as in the monologue experiment. As indicated in Table~\ref{tab:accu_mono_dialog} and~\ref{tab:corre_mono_dialog}, the results with conventional methods using the context (dialogue) show a decrease in performance compared to the monologue setting across most evaluation metrics. We hypothesize that merely concatenating utterances between two speakers is not an effective method for modeling the interactions between interlocutors. Therefore, we propose independently modeling both the interdependency among speakers and the intra-dependency within the speaker. The results, as shown in Tables~\ref{tab:accu_mono_dialog} and~\ref{tab:corre_mono_dialog}, indicate that our proposed method surpasses all baseline methods in the dialogue setting and marginally improves upon the results in the monologue setting.

\noindent\textbf{Data Augmentation in Dialogue.}
We test the effectiveness of data augmentation in the dialogue setting. The results, as shown in Table~\ref{tab:accu_dialog} and~\ref{tab:corr_dialog}, indicate that increasing speaker variety can enhance personality recognition in dialogue. Although the highest balanced accuracy achieved in the dialogue setting is 58.6, falling short of the monologue setting's best result of 60.4. Due to our focus on predicting only \textit{speaker a}'s personality in the dialogue setting, the original dataset lost half of its conversational data for augmentation purposes. This loss is an inevitable trade-off in the pursuit of speaker-independent personality recognition within dialogue settings.

\begin{table*}[t]
\caption{Accuracy results of HC-GNN in dialogue setting with various data sizes.}
\label{tab:accu_dialog}
\centering
\begin{tabular}{@{}llcccccc|cccccc@{}}
\toprule
\multicolumn{2}{c}{\multirow{2}{*}{\;\;\;\,Data}}      & \multicolumn{6}{c|}{Accuracy}      & \multicolumn{6}{c}{Balanced Accuracy}               \\ \cmidrule(l){3-14} 
\multicolumn{2}{l}{}           & N     & E     & O     & A     & C     & Avg. & N     & E     & O     & A     & C     & Avg. \\ \midrule
Monologue&+500k&62.7&58.2&\textbf{62.0}&57.9&\textbf{65.4}&61.2&\textbf{64.6}&\textbf{56.0}&\textbf{61.3}&60.3&\textbf{59.7}&\textbf{60.4}\\\midrule
\multirow{5}{*}{Dialogue} & Original&69.0&53.5&55.6&65.9&52.2&59.2&60.9&54.3&52.6&59.6&54.6&56.4\\
\multirow{5}{*}{(HC-GNN)}&+10k&73.9&59.7&54.0&63.9&54.6&61.2&59.5&52.8&50.7&55.6&59.0&55.5\\
&+20k&\textbf{74.3}&\textbf{60.1}&58.4&\textbf{67.5}&40.8&60.2&60.8&54.4&55.8&61.6&55.5&57.6\\
&+50k&69.6&54.7&58.7&62.2&63.8&\textbf{61.8}&61.9&55.5&58.4&\textbf{62.3}&55.1&58.6\\
&+500k&66.3&56.6&58.8&60.1&59.0&60.2&63.7&53.8&57.0&57.7&59.2&58.3\\ \bottomrule
\end{tabular}
\end{table*}

\begin{table}[t]
\caption{Correlation results of HC-GNN in dialogue setting with various data sizes.}
\centering
\label{tab:corr_dialog}
\begin{tabular}{@{}llccccc|ccccc@{}}
\toprule
\multicolumn{2}{c}{\multirow{2}{*}{\;\;\;\,Data}}      & \multicolumn{5}{c|}{Pearson Correlation}      & \multicolumn{5}{c}{Spearman Correlation}               \\ \cmidrule(l){3-12} 
\multicolumn{2}{l}{}           & N     & E     & O     & A     & C     & N     & E     & O     & A     & C     \\ \midrule
Monologue&+500k                            & \textit{.267} & .025  & \textbf{\textit{.459}} & \textit{.232} & \textit{.164}  & \textit{.333} & .045  & \textbf{\textit{.388}} & \textit{.230} & \textit{.203} \\\midrule          
\multirow{5}{*}{Dialogue} &Original & \textit{.285} & .040 & \textit{.347} & \textit{.216} & \textit{.169} & \textit{.304} & \textit{.066} & \textit{.243} & \textit{.191} & \textit{.193} \\
\multirow{5}{*}{(HC-GNN)}&+10k & \textit{.369} & .019 & \textit{.329} & \textit{.235} & \textbf{\textit{.234}} & \textit{.343} & .058 & \textit{.226} & \textit{.234} & \textbf{\textit{.263}} \\
&+20k & \textit{.327} & -.052 & \textit{.441} & \textbf{\textit{.314}} & \textit{.110} & \textit{.301} & .003 & \textit{.245} & \textit{.278} & \textit{.142} \\
&+50k & \textit{.345} & .030 & \textit{.312} & \textit{.300} & \textit{.162} & \textit{.374} & .050 & \textit{.250} & \textbf{\textit{.281}} & \textit{.183} \\
&+500k & \textbf{\textit{.426}} & .046 & \textit{.411} & \textit{.209} & \textit{.223} & \textbf{\textit{.439}} & .062 & \textit{.315} & \textit{.194} & \textit{.242} \\
\bottomrule 
\multicolumn{12}{r}{(\textit{Italic} means $p<.05$)}
\end{tabular}
\end{table}

\section{Conclusion}
We have proposed a data augmentation method for personality recognition, which involves interpolating between two existing data points to enhance speaker diversity. Additionally, we have introduced the HC-GNN method to independently model the interdependencies among interlocutors, as well as the intra-dependencies within the speaker in dialogues. Experimental results from the RealPersonaChat corpus demonstrate that increasing speaker diversity significantly improves personality recognition in both monologue and dialogue settings. Our HC-GNN method outperforms baseline models, showcasing its effectiveness. However, our experiments suggest that context did not make a large improvement in personality recognition. Further exploration of the dialogue setting will be the focus of our future work.

\begin{acknowledgement}
This work was supported by JST Moonshot R\&D Goal 1 Avatar Symbiotic Society Project (JPMJMS2011). This work was also supported by JST, the establishment of university fellowships towards the creation of science and technology innovation, Grant Number JPMJFS2123, and JSPS KAKENHI Grant Number 21J23124.
\end{acknowledgement}
\bibliographystyle{unsrtnat}
\bibliography{reference}

\end{document}